\documentclass[uspaper, 10 pt, conference]{format/ieeeconf} 

\IEEEoverridecommandlockouts                              

\usepackage{graphicx}
\usepackage[table,dvipsnames]{xcolor} 
\usepackage{amsmath,amssymb,amsfonts}
\usepackage{mathtools}                
\usepackage{algorithm}
\usepackage[noend]{algpseudocode}     
\usepackage{array,booktabs,multirow,eqparbox}
\usepackage[caption=false,font=footnotesize]{subfig}
\usepackage{tikz}
\usepackage{url,csquotes,pifont,tablefootnote,textcomp,scalerel}
\usepackage{siunitx}
\usepackage{balance}

\newcommand\copyrighttext{%
  \scriptsize\centering
  \textcopyright\ 2025 IEEE. Personal use of this material is permitted.
  Permission from IEEE must be obtained for all other uses, in any current or future
  media, including reprinting/republishing this material for advertising or promotional
  purposes, creating new collective works, for resale or redistribution to servers or
  lists, or reuse of any copyrighted component of this work in other works.\\
  Accepted manuscript. Published in the 2025 IEEE 21st International Conference on
  Automation Science and Engineering (CASE), pp. 2194--2199.
  DOI: \url{https://doi.org/10.1109/CASE58245.2025.11163761}.}
\newcommand\copyrightnotice{%
\enlargethispage{-0.7in}%
\begin{tikzpicture}[remember picture,overlay]
\node[anchor=south,yshift=10pt] at (current page.south) {\fbox{\parbox{\dimexpr\textwidth-\fboxsep-\fboxrule\relax}{\copyrighttext}}};
\end{tikzpicture}%
}
\renewcommand{\arraystretch}{1.3}
\graphicspath{ {./figures/} }
\title{\LARGE \bf
Isaac Sim-to-Real: Reinforcement Learning based Locomotion for Quadrupeds
}

\author{Jordan Dowdy$^{1}$ and Jean Chagas Vaz$^{2^*}$
\thanks{$^{1}$J. Dowdy is a Ph.D student with the Department of Electrical and Computer Engineering, University of Louisville. {\tt\small jordan.dowdy@louisville.edu}}
\thanks{$^{2^*}$Dr. Jean Chagas Vaz is with the Faculty of Electrical and Computer Engineering, University of Louisville, Louisville, KY, 40208, USA. $^{*}$direct all correspondence to this author {\tt\small jean.chagasvaz@louisville.edu}}
} 

\begin{document}
\maketitle
\thispagestyle{empty}
\pagestyle{empty}
\copyrightnotice
\begin{abstract}
Learning-based approaches to locomotion have risen in popularity in recent years, showing the capability for complex legged locomotion and whole-body control. Reinforcement learning (RL), the primary learning-based approach for locomotion, often utilizes a high-performance simulation tool, providing a controlled and efficient training and development environment. However, policies that perform well in simulation frequently encounter unexpected challenges when deployed on a physical system, known as the sim-to-real gap. This work presents a robust RL locomotion framework capable of whole-body control. The proposed RL framework utilizes Nvidia's new set of simulation tools, Isaac Sim, and its companion RL framework, Isaac Lab, for training, achieving a zero-shot sim-to-real policy. The performance of our policy is validated on physical hardware using the Unitree Go1, with experimental results showing similar velocity tracking performance to the quadrupeds' integrated controller, with a greater ability to recover from large disturbances, and achieve linear velocities of $2.0$ $m/s$ and angular velocities of $1.8$ $rad/s$.
\end{abstract}

\section{INTRODUCTION}

Quadrupedal locomotion is a key area of focus in legged robotics, with it chasing agile and robust mobility across various environments and terrains. Traditional locomotion control often combines high-level gait generation with stability control schemes such as through zero-moment point (ZMP) or model-predictive control (MPC) \cite{CheetahMPC, 8834241, 7970144}, with low-level high-frequency actuator controllers to track joint positions. While this hybrid approach can ensure basic mobility, it frequently struggles with the intricate, nonlinear dynamics encountered in irregular terrains. In contrast, learning-based strategies, specifically reinforcement learning (RL) techniques, leveraging high-performance simulation environments, have proven effective in developing robust and adaptive control policies \cite{2203.05194, 1901.08652}. 
In recent years, research for these RL locomotion techniques has seen a paradigm shift from high-level, gait generation and foot placement, to an all-encompassing approach, training policies to learn all required components of locomotion, and enabling whole-body control.

This demonstration of whole-body control potential in RL, has propelled industry leaders like OpenAI \cite{OpenAIgym}, NVIDIA \cite{IsaacGym}, and Google \cite{2502.08844} to each develop simulation platforms (OpenAI Gym, Isaac Sim, Mujoco), with accurate and optimized rigid-body dynamics \cite{2103.04616,1402.7050,9386154}. Using high-performance simulation environments for training, RL-based approaches provide a controlled and efficient way to develop complex control strategies. However, despite encouraging simulation results, the well-known sim-to-real gap presents significant challenges when implementing these policies on physical robots \cite{9606868}. Bridging this gap is critical to ensure the reliability and robustness of locomotion controllers in practical applications.
\begin{figure}[t]
    \centering
    \setlength{\fboxsep}{0pt}%
    \setlength{\fboxrule}{1.5pt}%
    \fbox{\includegraphics[width=\dimexpr\linewidth-2\fboxrule\relax]{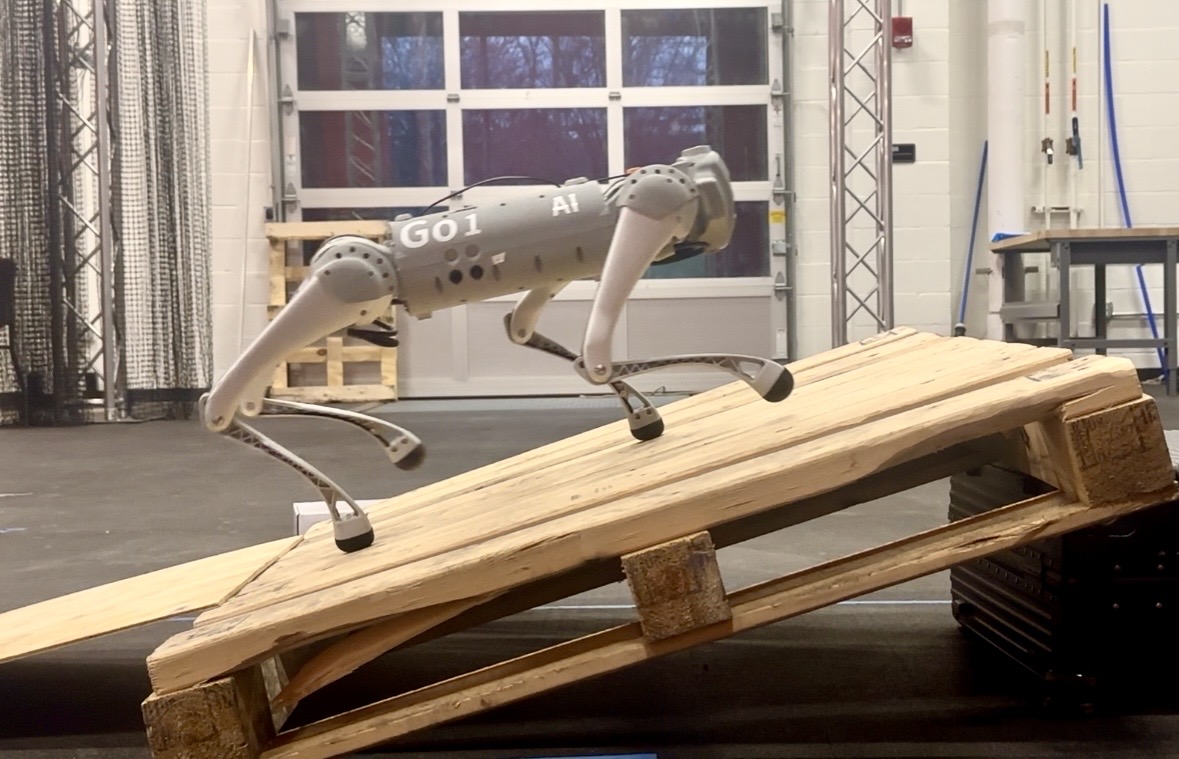}}
    \caption{\textbf{Unitree Go1 walking up an incline pallet} using a end-to-end RL locomotion framework.}
    \label{fig:Go1incline}
\end{figure}

\subsection{Literature Review}

Recent advancements in reinforcement learning (RL) for quadruped locomotion frequently result in policies that function as high-level controllers, achieving stable and optimal gaits \cite{yu2021visuallocomotion, 2110.15344}. Some studies integrate high-bandwidth sensors into the policies state to enhance locomotion in adverse terrains. These approaches still maintain the multi-modal approach \cite{10611128}. In contrast, other RL policies designed for whole-body control depend on pre-trained models within a teacher-student network architecture \cite{10605235}. The RL policies presented in this work share similarities with these approaches, while further incorporating advanced reward shaping techniques \cite{10886438} and employing a dual-policy framework to coordinate both high-level and low-level control \cite{10160760}.

While simulation significantly reduces training time and obviates the need for hardware \cite{IsaacLab}, several studies have demonstrated the effectiveness of real-world training and fine-tuning for locomotion policies \cite{10650813, 9812166}. However, simulation training with domain randomization, noise injection, and system disturbances \cite{10343164}, reduces the chance of encountering challenges and hitting the so-called sim-to-real gap. These domain randomization techniques, along with proper actuator modeling \cite{2212.03238, 2312.17507}, often prove sufficient to achieve real-world performance. Nvidia's Isaac Sim has seen a growing usage in recent years \cite{2407.08590}, along with its performance and capabilities. The latest release of Nvidia's RL framework, Isaac Lab, and physics solver, Physx5.4, introduces a suite of new and advanced tools, giving RL new potential. However, the transfer from simulation to physical hardware remains challenging despite significant improvements to these simulation environments.

To address this critical gap and build upon previous studies, we explore a robust RL-based locomotion framework, tailored to the Isaac Sim platform. Our RL locomotion framework produces a low-level position-control policy capable of whole-body control. Our policy does not utilize pre-acquired data as in teacher-student frameworks, and instead focuses on reward shaping to produce natural gaits during locomotion. The policy is subsequently validated on the Unitree Go1 quadruped through extensive hardware experiments, a performance benchmark against the platform's integrated controller.

\subsection{Paper Contributions}
\vspace{3pt}
\begin{itemize}

  \item An end-to-end RL-based locomotion policy with a zero-shot transfer to the Unitree Go1. 

  \item Techniques in domain randomization and actuator modeling to overcome the sim-to-real gap. 

  \item Policy validation and comparisons against Unitree's Go1 integrated controller, with comparable velocity tracking performance while achieving higher velocities. Policy disturbance rejection capabilities are also reported. 
\end{itemize}

\section{Reinforcement Learning Framework}
This section presents the methodology for training our RL locomotion framework. It starts with a formal definition of the problem, followed by a discussion of the approaches used to overcome the sim-to-real gap. Lastly, details on policy actions, observations, and rewards are discussed, along with the intuition behind them.

\begin{figure}[t!]
    \centering
    \setlength{\fboxsep}{0pt}%
    \setlength{\fboxrule}{1.5pt}%
    \fbox{\includegraphics[width=0.98\linewidth]{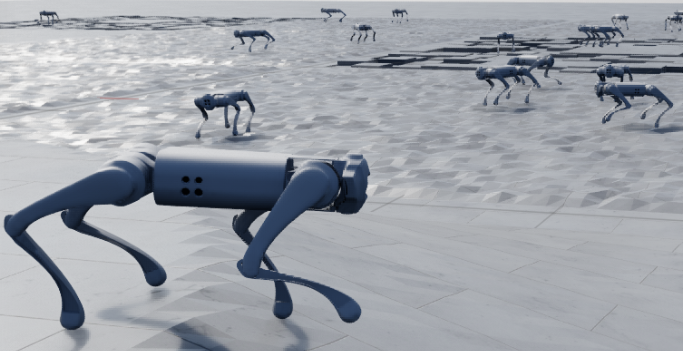}}
    \caption{\textbf{Simulation of Unitree Go1 in Isaac Sim} during policy inference. The simulation uses a terrain environment consisting of ramps, random height boxes, and a bumpy floor. During training, the terrain difficulty level as velocity tracking performance increases.}
    \label{fig:simEnviroment}
\end{figure}
\subsection{Problem Definition}
In reinforcement learning for the locomotion of a quadruped, a Markov Decision Process (MDP) is used to model how a robot interacts with its environment. A Markov Decision Process consists of ($\mathcal{S}, \mathcal{A}, \mathcal{P}, \mathcal{R}$, $\gamma$ ), where $\mathcal{S}$ and $\mathcal{A}$ are the current state and possible future state representing the agent's state space. $\mathcal{P}$ is the transition probability which shows the relationship between actions $\mathcal{A}_t$ and the change in states at $\mathcal{S}_t$ to the next time step $\mathcal{S}_{t+1}$, $\mathcal{R}$ is the reward function representing the reward received after taking action $\mathcal{A}_t$ in state $\mathcal{S}_t$, finally, $\gamma$ represents the discount factor and determines the importance of future rewards. The agent's goal is to maximize the reward given by: 
\begin{equation}
    G_t = \sum^{\infty}_{k=0} \gamma^k \mathcal{R}(\mathcal{S}_{t+k},\mathcal{A}_{t+k}),\label{eq:CDR}
\end{equation}
Where a policy $\pi(\mathcal{A}|\mathcal{S}$ defines the probability of taking an action in a state. The expected return starting from a state $\mathcal{S}$ under policy $\pi$ is formulated as:
\begin{equation}
    V_{\pi} = \mathbb{E}_{\pi} \sum^{\infty}_{k=0} \gamma^k \mathcal{R}(\mathcal{S}_{t},\mathcal{A}_{t}|\mathcal{S}_0=\mathcal{S}),\label{eq:SVF}
\end{equation}
Similarly, the expected return starting from a state $\mathcal{S}$ taking action $\mathcal{A}$ under policy $\pi$ is formulated as: 

\begin{equation}
    Q_{\pi}(\mathcal{S},\mathcal{A}) = \mathbb{E}_{\pi} \sum^{\infty}_{k=0} \gamma^k \mathcal{R}(\mathcal{S}_{t},\mathcal{A}_{t}|\mathcal{S}_0=\mathcal{S},\mathcal{A}_0=\mathcal{A})\label{eq:AVF},
\end{equation}
Where the optimal policy $\pi^*$ maximizes the following value functions,
\begin{equation}
    \pi^* = \arg \max_{\pi}V^{\pi}(\mathcal{S}),
\end{equation}
\begin{equation}
    \pi^* = \arg \max_{\pi}Q^{\pi}(\mathcal{S},\mathcal{A}).
\end{equation}

\subsection{Sim-to-Real Techniques}
This subsection details the techniques used to overcome the sim-to-real gap and achieve a zero-shot transfer. Additionally, implementation details are specified for domain randomization and actuator modeling for simulation. 

\subsubsection{Domain Randomization}
The NVIDIA RTX 3090 GPU was used to run 4096 environments simultaneously for policy training, utilizing domain randomization techniques during the training process, to ensure a robust policy and overcome the sim-to-real gap. The quantities foot friction, foot restitution, body mass, initial joint positions, and initial velocities were randomized to a difference of $\pm[5,20]\%$ their initial values. A terrain curriculum was used, gradually increasing terrain height and roughness during training, which can be seen in Fig.~\ref{fig:simEnviroment}. The terrain used during simulation creates boxes with slight variations in height close to 0 on the z-axis and bumpy patches by adding noise to terrain heights. Random pushing events were applied to the robot's base to disturb the system. These random pushing events would occur every 10 to 15 seconds by setting the linear velocity to a value in the range $[\text{-}1.0, 1.0]$~$m/s$. Command velocities are sampled from the ranges $[\text{-}2, 3]$~$m/s$ in x, $[\text{-}1.5, 1.5]$~$m/s$ in y, and $[\text{-}1.5, 1.5]$~$rad/s$ for linear and angular velocities, respectively, sampled every 0.3 to 10.0 seconds. Table \ref{tab:DomRand} provides more detail for all domain randomization quantities used. 

\begin{table}\centering
\caption{Domain Randomization}\label{tab:DomRand}
\setlength{\tabcolsep}{4pt}
\renewcommand{\arraystretch}{1.5}
\begin{tabular}{cccl}
\toprule
\textbf{Term}              & \textbf{Operation} & \textbf{Range}\\ 
\midrule
Foot Friction  &  New    & $(0.4, 1.1)$
\\
Center of Mass    & Add   & $(\text{-}0.025,0.025)$                    \\
Link Mass  & Scale        & $(0.8,1.3)$         \\
Body Velocity & Add & $(\text{-}\hspace{0.1em}0.25, 0.25)$ \\
Body Position & Add & $(\text{-}\hspace{0.1em}0.5,0.5)$ \\
Body Orientation & Add & $(\text{-}\hspace{0.1em}0.02,0.02)$ \\
Joint Positions  & Scale  & $(\text{-}\hspace{0.1em}0.3, 0.3)$ \\
Joint Velocities  & Scale  & $(\text{-}\hspace{0.1em}2.5, 2.5)$ \\
\bottomrule
\end{tabular}    
\end{table}

\subsubsection{Actuator Modeling}
In order to accurately model the actuators on the Unitree Go1, a multilayer perception (MLP) network was used to capture the non-linear dynamics and is often referred to as an actuator-net. The model created uses the same approach shown in \cite{2212.03238}. Joint information is recorded during the deployment of a policy and then used to train an MLP network to minimize the error between torque outputs based on a history of joint positions and velocities. The network trained on this data uses three layers of 64 neurons using the soft-sign activation function. The actuator-net was modeled with a $k_p=20$ and a $k_d=0.5$, providing a balance between joint stiffness and compliance during locomotion.
. Training the policy before integrating an actuator-net showed significant challenges with the sim-to-real gap.

\subsection{Action Space}
The policy's action space is represented by a 12$\times$1 vector corresponding to the joint positions of each actuator on the Unitree Go1 quadruped. These joint positions are scaled by a constant scale factor of $0.2$ and offset by the quadruped's nominal configuration. Scaling and offsetting these joint actions ensure that the robot has an initial stability during training, as it will make small initial movements around the nominal standing configuration. During training, these scaled and offset joint positions are then given to an actuator-net to correctly model output torques from the quadruped. 

\subsection{Observations}
The observation space for our policy takes a similar approach to past studies, utilizing minimal state information, represented by a 48$\times$1. The linear velocity, angular velocity, projected gravity, user command, joint positions, joint velocities, and previous action, are used as the policy states. Table~\ref{tab:obsTerms} gives a detailed look into these observations. The states related to IMU sensors (linear velocity, angular velocity, projected gravity) help the agent track desired velocity and model system disturbances due to foot slippage, collisions, and environment-applied forces. The user command state represents the desired linear (X, Y) and angular (Z) velocities that the policy needs to track. The joint position observation uses a relative position where $\theta_i$ is the nominal stance, encouraging the policy to stay at this configuration. The previous action term, given in its unscaled form, ensures the policy can learn how its actions change its subsequent states during locomotion. It should be noted that our states do not have any scaling applied, unlike other studies on RL locomotion. Scaling state information can help with policy converging to stable solutions if certain states are orders of magnitude larger.

\begin{table}\centering
\caption{Observation Space Terms}\label{tab:obsTerms}
\setlength{\tabcolsep}{4pt}
\renewcommand{\arraystretch}{1.3}
\begin{tabular}{cccl}
\toprule
\textbf{Observation Term}               & \textbf{State Expression}         & \textbf{Injected Noise}    \\
\midrule
Linear Velocity    &$ \upsilon_{base} \in \mathbb{R}^3$           &0.02                      \\   
Angular Velocity       & $\boldsymbol{\omega}_{yaw}$ $\in \mathbb{R}^3$    &0.01                       \\
Projected Gravity       & $\textbf{G}_{base} \in \mathbb{R}^3$    &0.05                      \\
User Command          &$ \textbf{u}^*_{base}\in\mathbb{R}^3$    &0.0                       \\
Joint Positions         & \textbf{$\left|\left|\theta_i-\theta\right|\right|$} $\in \mathbb{R}^{12}$   &0.05                      \\
Joint Velocity         & \textbf{$\Dot{\theta}$} $\in \mathbb{R}^{12}$     &0.1                       \\
Previous Action      & \textbf{$\mathcal{A}_{t-1.}$} $\in \mathbb{R}^{12}$      &0.0                       \\
\bottomrule
\end{tabular}    
\end{table}

\subsection{Rewards}
Through reward function shaping, the policy can achieve a stable periodic gate, maintaining desired command velocities. The Gait reward term enables the policy to achieve a natural gait during training by rewarding synchronous and asynchronous foot contact. To ensure the policy properly lifts the agent's feet off the ground, large weights are set to the feet clearance reward and the foot airtime reward. Joint position and velocity limit penalties are added for safety. Previous training attempts saw the policy reach joint limits for a single timestep; to prevent this, a large penalty weight was added. Thigh and calf contact penalties improved stability from previous attempts, preventing calf links from scraping the ground when moving backwards. Compared to other work, higher weights were used for action smoothness and joint deviation, which showed performance improvements at higher command velocities. Additionally, a penalty term is added to the amount of force applied to the agent's feet, allowing for a softer contact and a lower risk of hardware damage. 

\begin{table}\centering
\caption{Reward and Penalty Terms}\label{tab:rwdTerms}
\setlength{\tabcolsep}{4pt}
\renewcommand{\arraystretch}{1.5}
\begin{tabular}{cccl}
\toprule
\textbf{Reward Term}                    & \textbf{General Expression}        & \textbf{Weight}       \\
\midrule
Feet Air Time                & $\sum^4_{i=1}(t_{\mathit{air},i} - 0.175)$          &5.0           \\
Linear Velocity Error        & $\mathit{exp}(-(||\upsilon_{x,y}^{\mathit{des}} - \upsilon_{x,y}^{\mathit{cur}}|| )$      &5.0          \\
Angular Velocity Error  &$\mathit{exp}(-(||\omega_{\mathit{z}}^{\mathit{des}} - \omega_{\mathit{z}}^{\mathit{cur}}||/2) )$         &5.0 \\
Gait                         & $\prod^2_{i=1}f_{\mathit{sync},i}\cdot\prod^4_{i=1}f_{\mathit{async},i}$                     &10.0  \\
Foot Clearance                & $\mathit{exp}(-\sum^4_{i=1}(f_{\mathit{z},i} - 0.08))$                              &5.0           \\
\midrule
\textbf{Penalty Term}                   & \textbf{General Expression}                                           & \textbf{Weight}       \\
\midrule
Action Smoothness                & $||\mathcal{A}_t - \mathcal{A}_{t\text{-}\scaleto{1\mathstrut}{5.5pt}}||$                  &-1.5         \\
Air Time Variance                & $\sum^{4,4}_{i=1,j=1}(t_{\mathit{air},i} - t_{\mathit{air},j})$             &-1.0           \\
Base Motion                      & $| \omega_x + \omega_y |$      &-2.0         \\
Base Orientation                 & $||\textbf{G}_{x,y}||$                           &-5.0           \\
Foot Slippage                    & $||\textbf{C}_{\mathit{foot}} \cdot \upsilon_{\mathit{foot}}||$                         &-1.0           \\
Foot Force                       & $\sum^{4}_{i=1}(\textbf{F}_{\mathit{foot,~i}})^2$ & -0.00002 \\
Joint Torque                     & $||\tau||$                          &-0.0005           \\
Joint Acceleration               & $||{\Ddot{\theta}}||$                           &-0.0002           \\
Joint Velocity                   & $||\dot{\theta}||$                           &-0.015           \\
Joint Deviation             & $\sum^{12}_{i=1}(\theta_{\mathit{i}}-\theta)$  &-0.75 \\
Joint Position Limit       & $\sum^{12}_{h=1}(\theta\geq\theta_{\mathit{max}})$  &-100.0 \\
Joint Velocity Limit       & $\sum^{12}_{h=1}(\dot{\theta}\geq\dot{\theta}_{\mathit{max}})$  &-10.0 \\
Thigh Contact                          & \textbf{$\textbf{C}_{\mathit{thigh}}$} $\in \mathbb{R}^4$  &-10.0 \\
Calf Contact                          & \textbf{$\textbf{C}_{\mathit{calf}}$} $\in \mathbb{R}^4$  &-1.0 \\
\bottomrule
\end{tabular}

\end{table}

\begin{figure}
    \centering
    \includegraphics[width=3.4in]{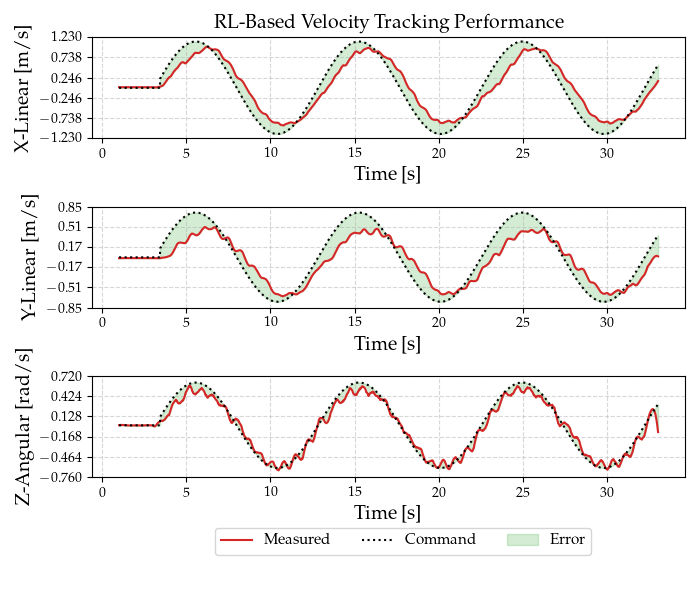}
    \caption{\textbf{Policy velocity tracking performance} with a generated sinusoidal velocity command. Each plot, from top to bottom, shows the measured, commanded, and error for linear velocities $x$,$y$, and yaw angular velocity in $z$.}
    \label{fig:velTrack}
\end{figure}

\begin{figure}
    \centering
    \includegraphics[width=3.4in]{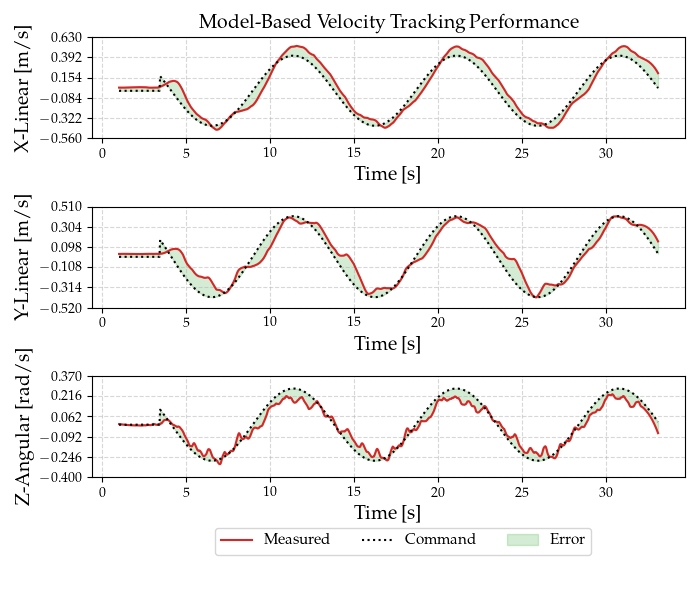}
    \caption{\textbf{Unitree's model-based velocity tracking performance} with a generated sinusoidal velocity command. Each plot, from top to bottom, shows the measured, commanded, and error for linear velocities $x$,$y$, and yaw angular velocity in $z$.}
    \label{fig:velmodelTrack}
\end{figure}

\begin{figure*}[!th]
    \centering
    \includegraphics[width=0.9\linewidth]{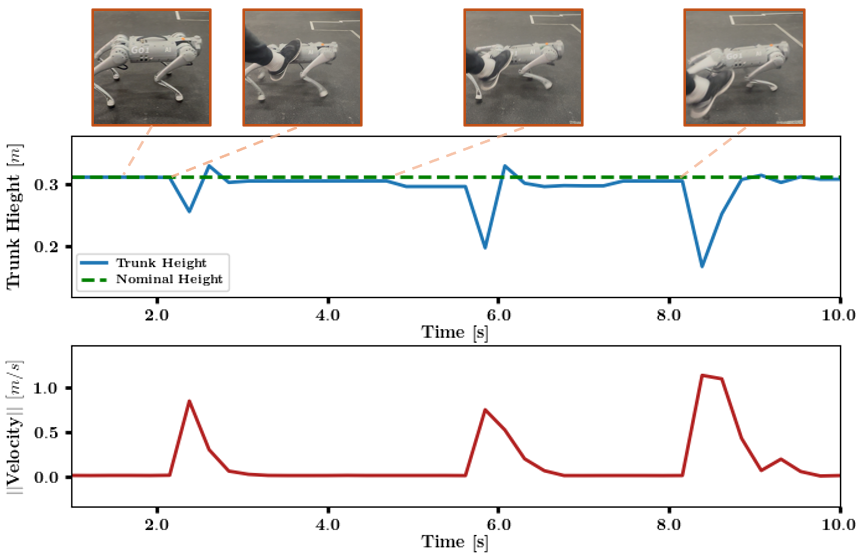}
    \caption{\textbf{Policy disturbance rejection} during three separate kicking events. Each kick increased in force applied to the trunk of the quadruped. The green line represents the quadruped's height in its nominal configuration, where, after a kicking event, the quadruped should return to this nominal height. The blue line represents the height of the quadruped during these events. As an approximation of the applied force from each kick, the linear velocity of the quadruped is shown by the red curve. The raw data was recorded at $25$$Hz$, with the policy running at $100$$Hz$.}
    \label{fig:push}
\end{figure*}

\section{Experimental Setup and Results}
With our RL-based locomotion framework now detailed, this section will focus on the network architecture used to deploy this policy. It will briefly summarize the network architecture and explain our experimental apparatus. Finally, real-world results showcase velocity tracking performance, model comparison, and the system's response to pushes.

\subsection{Network Architecture, Training, and Apparatus}
The reinforcement learning policy in this work is modeled using a straightforward actor-critic architecture using a multilayer perception network containing three different layers of 512, 256, and 128 neurons, employing exponential linear units (ELUs) as the activation functions, with both actor and critic networks using the same-sized MLP architecture. Proximal policy optimization (PPO) is used to train the policy with tuned hyper-parameters, including an adaptive learning rate, an entropy coefficient of $0.0025$, a target KL-Divergence at $0.01$, a discount factor ($\gamma$) and advantage estimator discount factor ($\lambda$) of $0.99$ and $0.95$ respectively, and a learning rate of $0.001$. The policy training is conducted over 5,000 iterations, taking approximately 3 hours to complete. The policy is trained with a physics $\Delta t$=$0.002$$s$, and a policy $\Delta t$=$0.02$$s$. During hardware testing of the trained policy, performance was seen to increase as the loop rate for the policy increased, going from its trained $50$$Hz$ to $100$$Hz$. This visually shows a more agile and stable performance during locomotion. 

The Unitree Go1 employs our end-to-end RL locomotion policy for evaluation. Unitree's Go1 is an 18-degree-of-freedom (DoF) quadruped with 12 actuated joints, expressing the trunk in a full 6 DoF. The Policy was compiled in ONNX format, with inference running on a Nvidia Jetson Nano—a Docker container with necessary Python dependencies to run the ONNX policy. Surprisingly, the policy and the necessary state collection and estimation can achieve up to $400Hz$ without GPU acceleration, simplifying the overall architecture. Linear velocity state information was determined through a Kalman Filter. To achieve a loop rate of over $75Hz$, the Kalman Filter was entirely implemented in C++, compiled using Pybind11 to create a Python library. This reduces latency enough to achieve up to $400Hz$ with the policy.

\subsection{Experimental Setup}
Experiments were conducted on physical hardware with the Unitree Go1 to study our learning based approach to locomotion, and compared against the integrated locomotion controller. The Unitree Go1 has three separate gaits, the default gait is used for general locomotion while the other two are specific for running and or stair climbing applications. To validate the locomotion policy, the Unitree Go1 was controlled to walk over rugged or adverse terrains such as gravel, sand, incline and decline plains. For comparison against our policy, the "default" gait is used, as this shows the general robustness and capabilities of the model-based locomotion controller. To show the tracking performance, a sinusoidal velocity command for linear $[x,y]$ and angular $[\omega_z]$ velocities were used and recorded for thirty-five seconds. As disturbance rejection is an essential part of quadrupedal locomotion, a performance test was performed to determine the disturbance rejection capabilities. The disturbance rejection test performed on the quadruped was enacted by kicking the trunk of the quadruped three separate times, simulating a fall or collision, and an externally applied force during locomotion. Each kicking event occurred consecutively, with each kick applying a higher level of force onto the trunk of the quadruped.

\subsection{Results}

Validation showed the capabilities to walk on various terrain difficulties with Fig.~\ref{fig:Go1incline}, showing the quadruped walking up a steep incline pallet. The policy can also be seen walking on declined and bumpy terrain in a simulation environment, Fig.~\ref{fig:simEnviroment}. Velocity tracking performance for the policy is shown in Fig.~\ref{fig:velTrack}, with Unitree's integrated controller performance shown in Fig.~\ref{fig:velmodelTrack}; however, instabilities in the model-based approach were seen at higher frequencies and amplitude (large accelerations and velocities) for the sinusoidal command velocity, causing the feet to slip. To mitigate this issue, the maximum and minimum velocities were reduced. The policy tracking performance shows a small phase lag between the desired and measured velocities, with Unitree's model-based approach producing similar results. Our learning based approach is shown to match or produce similar results in comparison to a model-based approach. The results for the disturbance rejection abilities of the policy can be seen in Fig.~\ref{fig:push}. A velocity plot at the bottom provides a general estimate of the applied force on the trunk from the kick. Additionally, the quadruped's nominal height shows how well the policy recovered from the disturbance. The policy successfully recovered from all three kicks while only stumbling and never entirely falling to the ground. These results indicate that the policy is robust against events such as collisions or foot slippage during locomotion. 

\section{Conclusion}
In this paper, an end-to-end reinforcement learning locomotion framework with a zero-shot transfer was presented, showing the capability to outperform or match the capabilities of modern model-based approaches such as the controller integrated on Unitree's Go1. Additionally, the techniques used to overcome the sim-to-real gap for Isaac Sim, along with observation and reward shaping, were used to achieve these results. This RL framework showed its capability to estimate the joint angles needed to achieve robust locomotion with hardware results. The RL framework also showed the ability to closely track a desired velocity command from a user. This controller was capable of resisting and had the ability to recover from large pushing forces.

\nocite{*}
\bibliography{./format/IEEEabrv.bib, references}
\bibliographystyle{./format/IEEEtran.bst}
\end{document}